\documentclass[10pt,twocolumn,letterpaper]{article}

\usepackage{iccv}
\usepackage{times}
\usepackage{epsfig}
\usepackage{graphicx}
\usepackage{amsmath}
\usepackage[accsupp]{axessibility} 

\usepackage[pagebackref=true,breaklinks=true,letterpaper=true,colorlinks,bookmarks=false]{hyperref}

\usepackage{amsfonts,amsthm}
\usepackage{array}
\usepackage{authblk}
\usepackage{booktabs}
\usepackage{color}
\usepackage{enumitem}
\usepackage{float}
\usepackage{mathtools}
\usepackage{multirow}
\usepackage{setspace}
\usepackage{subcaption}

\theoremstyle{definition}

\newcolumntype{L}[1]{>{\raggedright\let\newline\\\arraybackslash\hspace{0pt}}m{#1}}
\newcolumntype{C}[1]{>{\centering\let\newline\\\arraybackslash\hspace{0pt}}m{#1}}
\newcolumntype{R}[1]{>{\raggedleft\let\newline\\\arraybackslash\hspace{0pt}}m{#1}}

\setlist[itemize]{noitemsep, topsep=0pt}
\setlist[enumerate]{noitemsep, topsep=0pt}

\newcommand{\parens}[1]{\left(#1\right)}
\newcommand{\braces}[1]{\left\{#1\right\}}
\newcommand{\bracks}[1]{\left[#1\right]}

\newcommand{\norm}[1]{\left\Vert#1\right\Vert}

\iccvfinalcopy 

\def\iccvPaperID{5451} 

\ificcvfinal\fi

\begin{document}

\title{HighlightMe: Detecting Highlights from Human-Centric Videos}


\author[1]{Uttaran Bhattacharya\thanks{Work done while Uttaran an intern at Adobe Research}}
\author[2]{Gang Wu}
\author[2]{Stefano Petrangeli}
\author[2]{Viswanathan Swaminathan}
\author[1]{Dinesh Manocha}
\affil[1]{University of Maryland, College Park, MD, USA. \qquad {\tt\small \{uttaranb|dmanocha\}@umd.edu}}
\affil[2]{Adobe Research, San Jose, CA, USA. \quad\qquad\qquad {\tt\small \{gawu|petrange|vishy\}@adobe.com}}

\maketitle
\ificcvfinal\thispagestyle{empty}\fi

\begin{abstract}
    We present a domain- and user-preference-agnostic approach to detect highlightable excerpts from human-centric videos. Our method works on the graph-based representation of multiple observable human-centric modalities in the videos, such as poses and faces. We use an autoencoder network equipped with spatial-temporal graph convolutions to detect human activities and interactions based on these modalities. We train our network to map the activity- and interaction-based latent structural representations of the different modalities to per-frame highlight scores based on the representativeness of the frames. We use these scores to compute which frames to highlight and stitch contiguous frames to produce the excerpts. We train our network on the large-scale AVA-Kinetics action dataset and evaluate it on four benchmark video highlight datasets: DSH, TVSum, PHD$^2$, and SumMe. We observe a 4--12\% improvement in the mean average precision of matching the human-annotated highlights over state-of-the-art methods in these datasets, without requiring any user-provided preferences or dataset-specific fine-tuning.
\end{abstract}

\section{Introduction}\label{sec:intro}
Human-centric videos focus on human activities, tasks, and emotions~\cite{human_centric_video_understanding,moviegraph}. These videos form a major part of the rapidly growing volume of online media~\cite{cisco}, coming from multiple \textit{domains}, such as amateur sports and performances, lectures, tutorials, video weblogs (vlogs), and individual or group activities, \textit{e.g.}, cookouts and holiday trips. However, unedited human-centric videos also tend to contain large chunks of irrelevant and uninteresting content, requiring them to be edited for efficient browsing~\cite{lsvm_dsh}.

\begin{figure}[t]
    \centering
    \includegraphics[width=0.92\columnwidth]{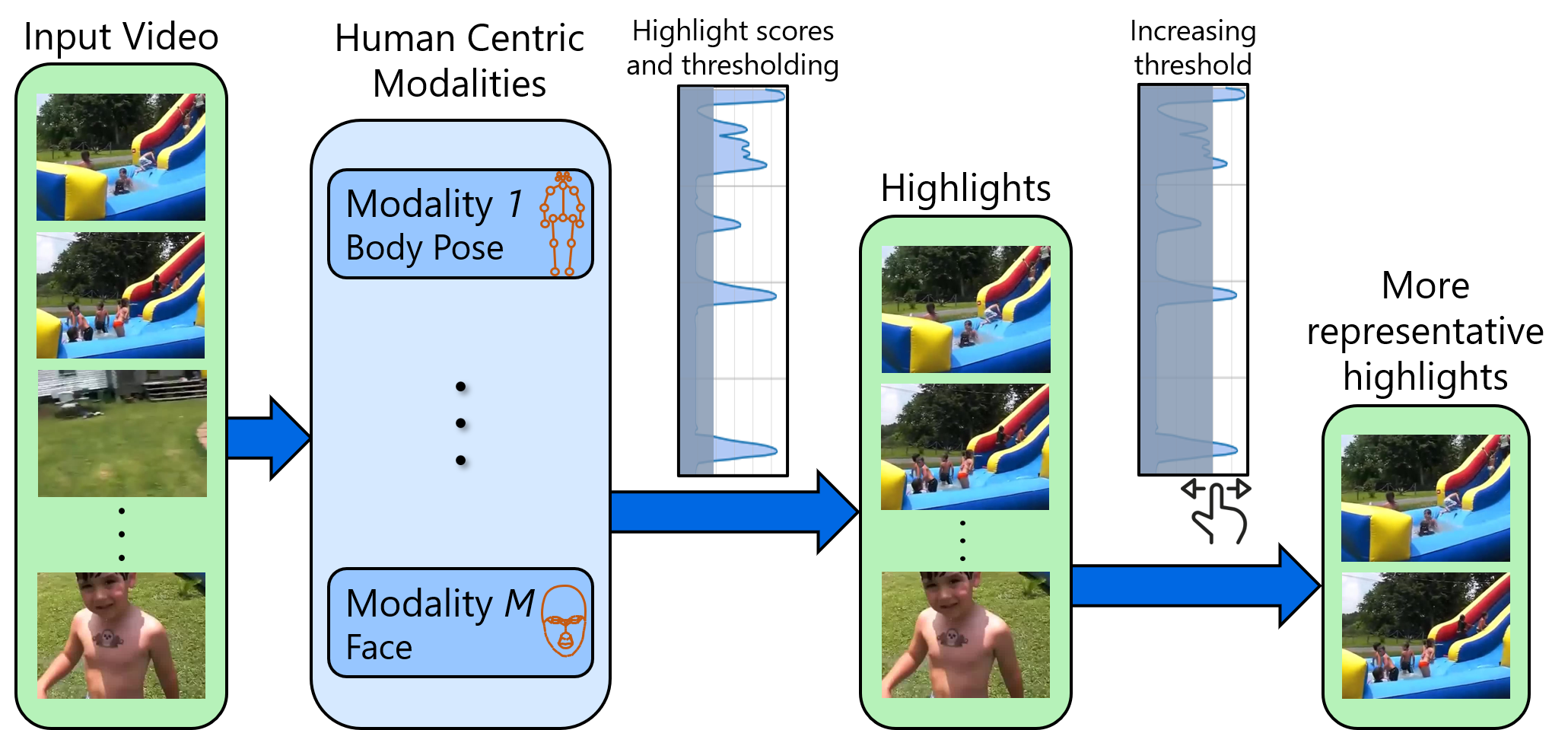}
    \caption{\textbf{Detecting highlight excerpts using human-centric modalities.} Our method leverages multiple human-centric modalities, \textit{e.g.}, body poses and faces, observable in videos focusing on human activities, to detect highlights. We use a 2D or 3D interconnected point representation of each modality to construct a spatial-temporal graph representation to compute the highlight scores.
    }
    \label{fig:teaser}
    \vspace{-15pt}
\end{figure}

To address this problem, researchers have developed multiple techniques for detecting highlightable excerpts and summarizing videos~\cite{phd2,less_is_more,adaptive_fcsn,sup_video_summ,fcsn,drl_summ}. Given unedited footage, highlight detection obtains the moments of interest, and summarization computes the most relevant and representative set of excerpts. Detecting effective highlights not only expedites browsing, but also improves the chances of those highlights being shared and recommended~\cite{less_is_more}. Current methods can learn to detect these excerpts given annotated highlights~\cite{lsvm_dsh,phd2}, or sets of exemplars for different highlight categories, \textit{e.g.}, learning from skiing images to detect skiing excerpts from videos~\cite{joint_summ,triplet_deep_ranking}. Other methods obviate the need for supervision by learning the representativeness of each frame or shot with respect to the original video~\cite{unsup_video_summ} and exploiting video metadata such as duration~\cite{less_is_more} and relevance of shots~\cite{drl_summ,retro_video_summ}. All these methods either assume or benefit from some domain-specific knowledge of the unedited footage, \textit{e.g.}, running and jumping may be more relevant in a parkour video, whereas sliding maneuvers may be more relevant in a skiing video. Alternative methods do not consider domain-specific knowledge but consider the pre-recorded preferences of multiple users instead to detect personalized highlights~\cite{adaptive_fcsn}.

Whether they assume domain-specific knowledge or user-preferences, existing methods work in the 2D image space of the frames or shots constituting the videos. State-of-the-art image-based networks can learn rich semantic features capturing the interrelations between the various detected objects in the images, leading to efficient highlight detection. However, these approaches do not explicitly model human activities or inter-person interactions that are the primary focus of human-centric videos. Developing methods for human-centric videos, meanwhile, has been essential for a variety of tasks, including expression and emotion recognition~\cite{deep_fexr,step,m3er}, activity recognition~\cite{stgcn}, scene understanding~\cite{moviegraph,zoom_in}, crowd analysis~\cite{panda}, video super-resolution~\cite{zoom_in}, and text-based video grounding~\cite{video_grounding}. These methods show that human-centric videos need to be treated separately from generic videos, by leveraging human-centric modalities such as poses and faces. Therefore, there is both the scope and the need to bring the machineries of human-centric video understanding to the task of highlight detection as well.

\textbf{Main contributions.}
We develop an end-to-end learning system that detects highlights from human-centric videos without requiring domain-specific knowledge, highlight annotations, or exemplars. Our approach utilizes the human activities and interactions that are expressed through multiple sensory channels or modalities, including faces, eyes, voices, body poses, and hand gestures~\cite{body_cues_not_faces,m3er}. We use graph-based representations for all the human-centric modalities to sufficiently represent how the inherent structure of each modality evolves with various activities and interactions over time. Our network learns from these graph-based representations using spatial-temporal graph convolutions and maps the per-frame modalities to \textit{highlight scores} using an autoencoder architecture. Our highlight scores are based on the representativeness of all the frames in the videos, and we stitch together contiguous frames to produce the final excerpts. Our novel contributions include:
\begin{itemize}
    \item \textbf{Highlight detection with human-centric modalities.} Our method identifies the observable modalities, such as poses and faces, in each input video and encodes their inter-relations, across both time and different persons, into \textit{highlight scores} for highlight detection.
    
    \item\textbf{Annotation-free training of highlight scores.} We do not require highlight annotations, exemplars, user-preferences, or domain-specific knowledge. Instead, we only need to detect of one or more human-centric modalities using off-the-shelf modality detection techniques to train our highlight scores.

    \item\textbf{Domain- and user-agnostic performance.} Our trained network achieves state-of-the-art performance in highlight detection over a diverse range of domains and user preferences, evaluated over multiple benchmark datasets consisting of human-centric videos.
\end{itemize}

Our method achieves a mean average precision of $0.64$ and $0.20$ of matching human-annotated highlight excerpts on the benchmark domain-specific video highlight (DSH) dataset~\cite{lsvm_dsh} and the personal highlight detection dataset (PHD$^2$)~\cite{phd2} dataset, respectively, and outperform the corresponding state-of-the-art methods by $7\%$ and $4\%$ (absolute). We also achieve state-of-the-art performance on the smaller benchmark datasets of TVSum~\cite{tvsum} and SumMe~\cite{summe}, outperforming the current state-of-the-art baselines by $12\%$ and $4\%$ (absolute) on the mean average precision and mean F-score, respectively. Even for domains that are not fully human-centric (\textit{e.g.}, dog shows) or videos where human-centric modalities are sparsely detected, the performance of our method is comparable to the current state-of-the-art.

\section{Related Work}\label{sec:rw}
Both highlight detection and the closely related problem of video summarization have been well-studied in computer vision, multimedia, and related fields. Early methods utilized a variety of techniques including visual-content-based clustering, scene transition graphs, temporal variance of frames~\cite{video_clustering,graph_modeling_video_summ,video_abstraction_survey}, and hand-crafted features representing semantic information such as facial activities~\cite{looking_at_viewer}. On the other hand, recent approaches have capitalized on an impressive range of deep learning tools and techniques to perform highlight detection and video summarization.

\textbf{Highlight Detection.} The goal of highlight detection is to detect interesting moments or excerpts from unedited videos~\cite{video_abstraction_survey,lsvm_dsh}. A large contingent of methods pose this as a supervised ranking problem, such that the highlightable excerpts are ranked higher than all other excerpts~\cite{lsvm_dsh,video2gif,pairwise_deep_ranking,att_deep_ranking,phd2,deep_ranking_360_video,region_based_deep_ranking,s2n}. These methods assume the availability of human-annotated labels of the highlightable excerpts and train networks to learn either generic or domain-specific ranking metrics that correlate with these labels. On the other hand, weakly-supervised and unsupervised highlight detection methods eliminate label dependencies by leveraging exemplars or video metadata. Exemplars include scraped web images depicting domain-specific actions such as gymnastics and skiing~\cite{triplet_deep_ranking}. Video metadata include information on video categories~\cite{rrae}, or properties useful in differentiating unedited videos from edited videos, \textit{e.g.}, duration~\cite{less_is_more}. Some approaches also take user preferences into account to generate personalized highlights~\cite{adaptive_fcsn}. All these methods perform computations in the 2D image space of the frames and do not utilize human-centric modalities.

\textbf{Video Summarization.} Video summarization aims to provide succinct synopses of videos in a variety of formats, including storyline graphs~\cite{storyline_graph_1,storyline_graph_2}, keyframe sequences~\cite{egocentric_video_summ}, clips~\cite{summe,retro_video_summ}, and their mixtures based on user requirements~\cite{rule_them_all}. It is commonly posed as a subsequence estimation task satisfying coherence~\cite{story_driven_summ}, diversity, and representativeness~\cite{collab_summ,drl_summ}. Existing unsupervised approaches build on multiple concepts, such as visual co-occurrence~\cite{mbf}, temporal relevance between frames and shots~\cite{joint_summ,unsup_video_summ,fcsn,retro_video_summ}, learning category-aware classifiers~\cite{category_specific_summ} and category-aware feature learning~\cite{quasi_real_time_summ,tvsum}. Weakly supervised approaches use exemplar web images and videos~\cite{storyline_graph_1,web_image_prior_summ,weakly_sup_summ,unpaired_data_video_summ}, and category descriptions~\cite{category_specific_summ,collab_summ} as priors. Other approaches use supervised learning with human-annotated summaries, using subset selection~\cite{diverse_sequential_subset}, visual importance scores~\cite{egocentric_video_summ,summe}, submodular mixtures~\cite{submodular_mixtures_1,submodular_mixtures_2}, and temporal inter-relations~\cite{sup_video_summ,retro_video_summ,hierarchical_rnn_summ}.
While our objective is highlight detection, our approach is inspired by these summarization methods. Particularly, we ensure that our highlight score captures the representativeness in the videos and satisfies robust feature reconstruction.

\textbf{Multimodal Learning.} A wide body of work has focused on multimodal action recognition~\cite{multimodal_dataset_action,deep_multimodal_action,skeleton_guided_action,skeleton_rgbd_multimodal_action} and emotion recognition~\cite{iemocap,audio_visual_emotion,cmu_mosei,m3er,emoticon}. These methods observe and combine cues from multiple modalities of human expression, including faces, poses, vocal tones, eye movements hand and body gestures, and gaits. Existing methods commonly model observed modalities using points and graphs~\cite{skeleton_guided_action,iemocap,m3er}, making them suitable for learning action- and emotion-specific features. In our work, we utilize the fact that highlightable excerpts of human-centric videos can be determined based on the modalities. Following recent trends in multimodal action and emotion recognition~\cite{skeleton_guided_action,m3er}, we also model the modalities observed across the frames in videos as spatial temporal graphs, and leverage them to learn our highlight scores.

\section{Multimodal Highlight Detection}\label{sec:multimodal_hlt}
Given human-centric videos, our goal is to detect the moments of interest or \textit{highlights} from the videos. This section elaborates on how we detect such highlights by leveraging the human-centric modalities observed from the videos.

\begin{figure}[t]
    \centering
    \includegraphics[width=0.88\columnwidth]{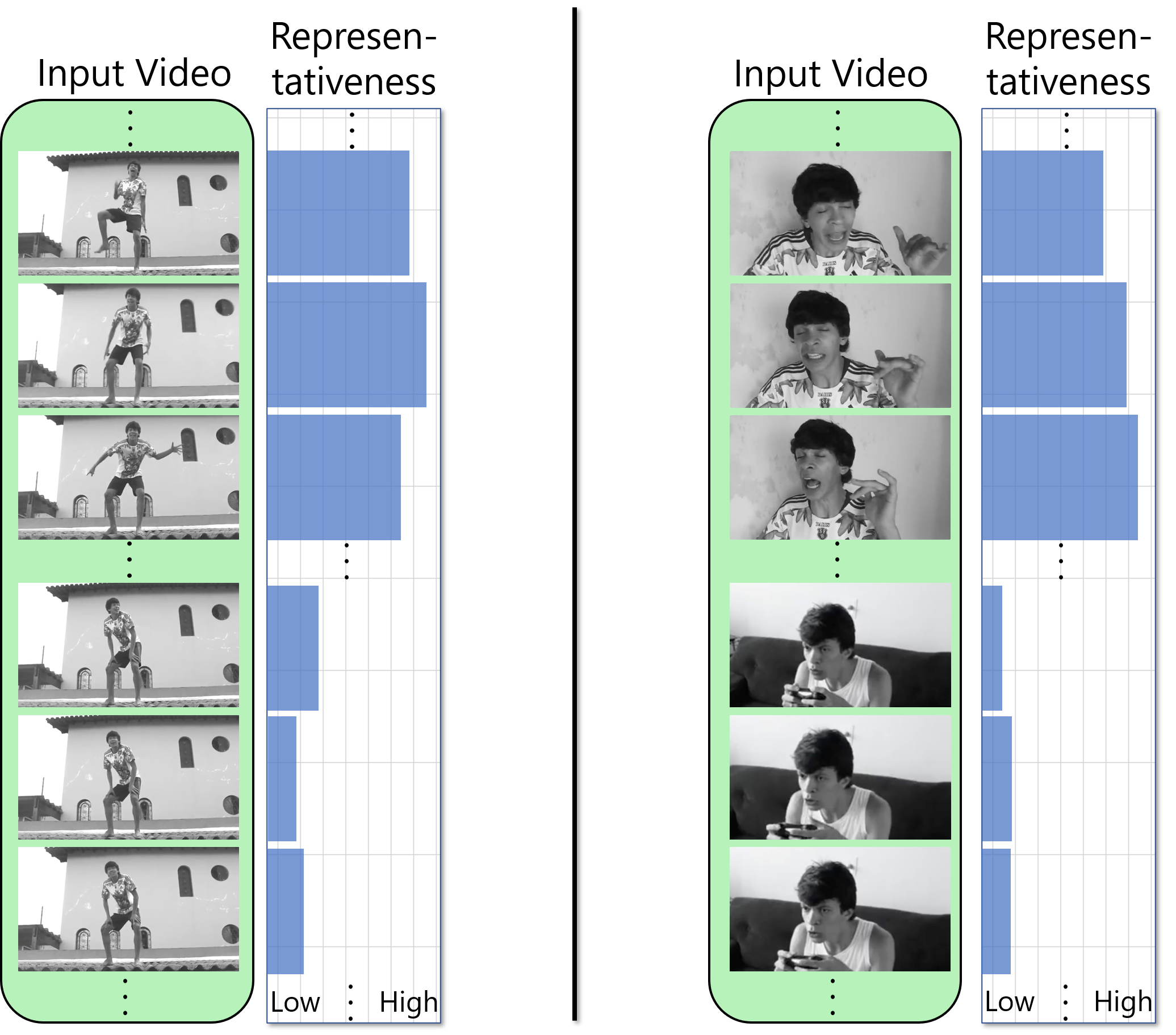}
    \caption{\textbf{Representativeness.} We show frames with different values of representativeness calculated in the space of poses \textit{(left)} and face landmarks \textit{(right)}. We learn highlight scores based on the representativeness.}
    \label{fig:representativeness}
    \vspace{-15pt}
\end{figure}

\subsection{Human-Centric Modalities}
In our work, we use the term \textit{modalities} to imply the channels of human expression sensitive to human activities and interactions, \textit{e.g.}, faces, eyes, body poses, hands, and gaits~\cite{multimodal_dataset_action,m3er,emoticon}. Activities constitute individual expressions and interactions occur with other humans, living beings, and inanimate objects, pertinent to a variety of actions~\cite{stgcn,skeleton_rgbd_multimodal_action} and emotions~\cite{step,emoticon}. We argue that the highlightable excerpts of human-centric videos preferred by human users focus on these activities and interactions. Therefore, we aim to learn from the observable human-centric modalities in our network. For each detected modality of each human, our network leverages the inter-relations at different time instances and the inter-relations between different humans to detect the most representative excerpts.

While we extract these modalities from the RGB image-space of the video frames, we note that the modalities better capture the rich semantics of the frames. Image-space representations build on variants of the intensity differences between different parts of images, without an underlying insight on how the different parts physically interact. Conversely, modalities provide insight on such interactions based on their structure, \textit{e.g.}, the relative movements of arms and legs indicate certain actions, and the relative movements of various facial landmarks indicate certain expressions and emotions. We build our network to explicitly consider the structure of each modality and the evolution of those structures with activities and interactions over time.

We consider $M \geq 1$ observable human-centric modalities from an input video. We assume the modalities are extracted using standard detection and tracking techniques~\cite{mpt,face_landmark_detect}, and are represented using a set of interconnected points in 2D or 3D, such as a set of 2D face landmarks for the face or a set of 3D body joints for the pose.

To represent each modality $m = 1, \dots, M$, we construct a spatial-temporal graph representation $\mathcal{G}_m = \braces{\mathcal{V}_m, \mathcal{E}_m}$. The nodes in $\mathcal{V}_m$ represent the points of the corresponding modality, and the edges in $\mathcal{E}_m$ represent both the structure of the modality and how that structure evolves over time. To sufficiently capture this, we consider three edges types:
\begin{itemize}
    \item \textbf{Intra-person edges} capturing the spatial relationships between the nodes of a single person, \textit{e.g.}, bones between pose joints and connectors between face landmarks. These edges represent the baseline structure of the modality at every video frame.
    \item \textbf{Inter-person edges} connecting the identical nodes of different persons, \textit{e.g.}, root to root, head to head, at every video frame. These edges capture how the nodes of different persons interact with each other. They form a bipartite graph for every pair of persons, and represent the inter-person interactions at every video frame.
    \item \textbf{Temporal edges} connecting the identical nodes of a person, \textit{e.g.}, root to root, head to head, over multiple video frames. These edges capture how those nodes evolve with time for every person. They form a bipartite graph for every pair of video frames, and represent the evolution of activities and interactions over time.
\end{itemize}
The spatial positions of these nodes and the combination of all these edges allow our network to learn the activities and interactions of all humans in videos and learn highlight scores accordingly, without any prior knowledge on the video domains or user-provided preferences.

\begin{figure}[t]
    \centering
    \includegraphics[width=\columnwidth]{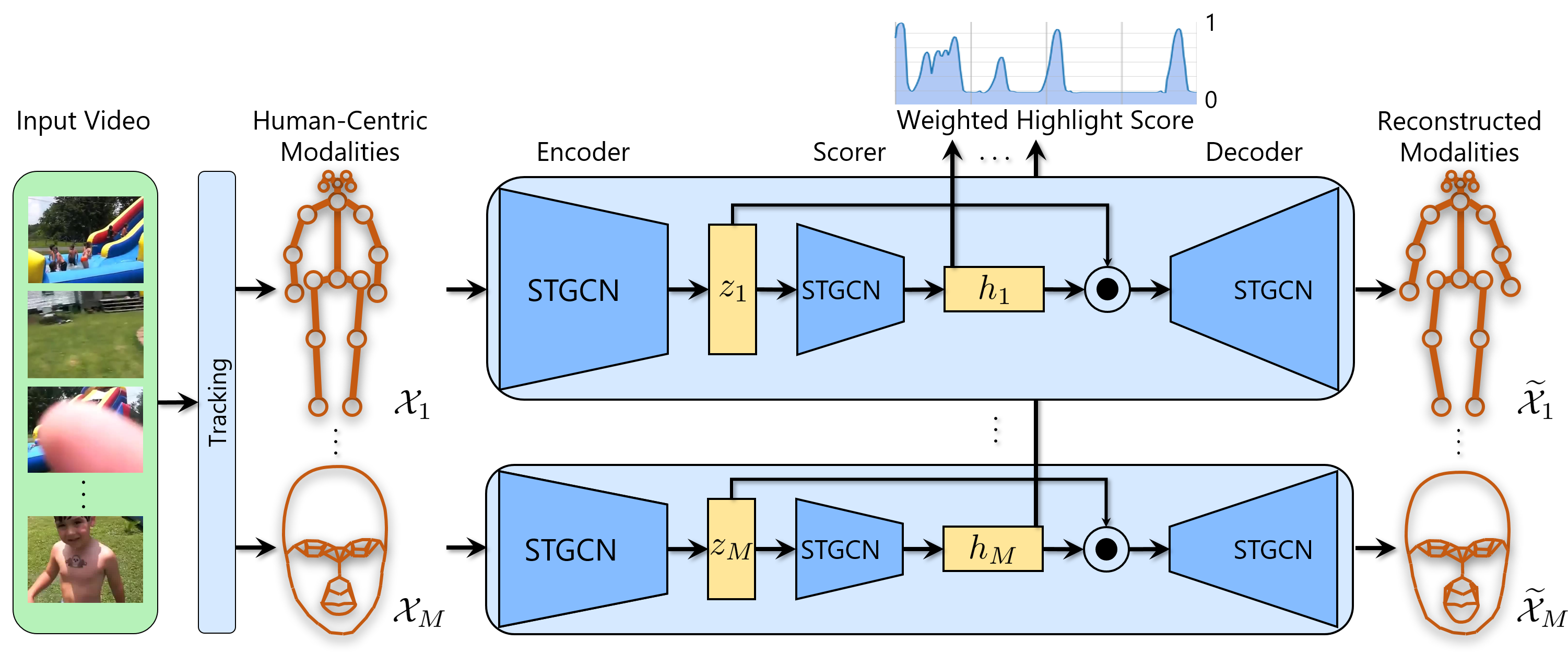}
    \caption{\textbf{Highlight detection with human-centric modalities:} Overview of our network for learning highlight scores from multiple human-centric modalities. We use standard techniques~\cite{mpt,face_landmark_detect} to detect the human-centric modalities. We represent the modalities as sets of connected points in either 2D or 3D. We train the networks for all the modalities in parallel. The only point of interaction between the networks is their predicted highlight scores, which we combine into our weighted highlight score for training.}
    \label{fig:unsup_net}
    \vspace{-15pt}
\end{figure}

\subsection{Representativeness of the Video Frames}\label{subsec:representativeness}
Since we aim to detect highlights from videos without requiring annotations or exemplars, our approach is aligned with detecting the representative frames from the videos, similar to video summarization~\cite{unsup_video_summ,rule_them_all}. While detecting the representative frames in the image-space may or may not lead to detecting the moments of interest for highlight detection~\cite{lsvm_dsh}, our key observation is that detecting the representative frames in the space of human-centric modalities, in fact, leads to detecting the moments of interest for highlight detection in human-centric videos.

We define the \textit{representativeness} of a video frame as the difference, in some metric space, between the video and the video without that frame. The larger the difference, the higher the representativeness of that frame. Intuitively, the representativeness of a frame measures the fraction of information it contains in relation to the entire video. Our goal in highlight detection is to detect a \textit{minimal} set of frames from a video with a \textit{maximal} representativeness.

In our work, we measure the representativeness in the metric space of the observable modalities. Figure~\ref{fig:representativeness} shows examples of frames with different values of representativeness in the space of poses and face landmarks. We consider each video to consist of a total $T$ frames and $P$ persons (zero-padding videos with fewer frames). Therefore, for each modality $m$, $\mathcal{V}_m$ consists of $N \times T \times P$ nodes in total, where $N$ is the number of nodes per person. We collate these nodes into a tensor $\mathcal{X}_m = \bracks{\mathcal{X}_m^{\parens{1}}; \dots; \mathcal{X}_m^{\parens{T}}}$, where $\mathcal{X}_m^{\parens{t}} \in \mathbb{R}^{N \times P \times D}$ for each frame $t$, and $D$ is the spatial dimension of each node, most commonly $2$ or $3$. We can then multiply a \textit{highlight score} $h_m^{\parens{t}}$ of $0$ or $1$ to each frame $t$ to reflect their representativeness. Thus, we can write the net difference $\mathcal{D}$ as a result of assigning the highlight scores as,
\begin{equation}
    \mathcal{D} = \norm{\mathcal{X}_m - \bracks{h_m^{\parens{1}}; \dots; h_m^{\parens{T}}} \odot \mathcal{X}_m},
    \label{eq:net_difference}
\end{equation}
where $\odot$ denotes the Hadamard product. We can now rewrite our goal as simultaneously minimizing $\mathcal{D}$ in Eq.~\ref{eq:net_difference} and $\sum_t h_m^{\parens{t}}$, for each modality $m$.

We note that a trivial solution to Eq.~\ref{eq:net_difference} is to pick a threshold $0 \leq \tau \leq T$, then assign a highlight score of $1$ to the top $\tau$ most representative frames from $\mathcal{X}_m$, and a highlight score of $0$ to all other frames. However, the choice $\tau$ is non-trivial and needs to be learned from the data in practice. Therefore, we train an autoencoder-based deep neural network to learn the highlight scores for a wide range of data. We also allow the highlight scores to be continuous in $\bracks{0, 1}$ to keep our network differentiable. Moreover, making the highlight scores continuous also helps us understand the relative representativeness of each frame, which is an inbuilt component of modern highlight detection systems~\cite{less_is_more,adaptive_fcsn}.

\subsection{Network Architecture}\label{subsec:arch}
Figure~\ref{fig:unsup_net} shows our overall network architecture for predicting highlight excerpts from input videos. The goal of our network is to learn per-frame highlight scores to minimize an analogous form of Eq.~\ref{eq:net_difference}. Our network achieves this by taking in the per-frame graph-based representations of the observable human-centric modalities. It attempts to reconstruct all the activities in the video using as few frames of the input modalities as possible, \textit{i.e.}, a weighted reconstruction, where the weights are the highlight scores. In this training process, our network learns to assign higher highlight scores to the frames with higher representativeness. We now describe our network architecture in detail.

Our autoencoder consists of an encoder, a scorer, and a decoder. Our encoder takes in the spatial-temporal graph $\mathcal{G}_m = \braces{\mathcal{V}_m, \mathcal{E}_m}$ for each observable modality $m$ from an input video. It uses a separate spatial temporal graph convolutional network (STGCN)~\cite{stgcn,gcn_1,gcn_2} to transform $\mathcal{X}_m$ of each modality $m$ into a latent activity-based feature $z_m \in \mathbb{R}^{N \times T \times P \times D_l}$, $D_l$ being the dimension of each node in the latent feature. We thus have the operation,
\begin{equation}
    \vspace{-5pt}
    z_m = \textrm{STGCN}\parens{\mathcal{A}_m, \mathcal{X}_m; W_m^{\parens{\textrm{enc}}}},
    \vspace{-2pt}
\end{equation}
where $\mathcal{A}_m$ denotes the adjacency matrix obtained from $\mathcal{E}_m$, and $W_m^{\parens{\textrm{enc}}}$ consists of the set of trainable STGCN parameters in the encoder. We note here that the data $\mathcal{X}_m$ forms a full-rank tensor, therefore the STGCN avoids the degenerate solution of assigning $0$'s to all $z_m$'s.

Our latent activity-based features $z_m$ $\forall m$ connect to our scorer, which consists of a single layer of spatial temporal graph convolution followed by a sigmoid operation per modality. Our scorer transforms each $z_m$ into a normalized highlight score $h_m \in \bracks{0, 1}^{N \times T \times P \times 1}$ for each node, \textit{i.e.},
\begin{equation}
    \vspace{-5pt}
    h_m = \sigma\parens{\textrm{STGCN}\parens{\mathcal{A}_m, z_m; W_m^{\parens{\textrm{hlt}}}}},
    \vspace{-5pt}
\end{equation}
where $\sigma\parens{\cdot}$ represents the sigmoid function and $W_m^{\parens{\textrm{hlt}}}$ consists of the set of trainable STGCN parameters.

Our decoder takes in the feature $z_m$ and the highlight score $h_m$ for each modality $m$, and produces a weighted latent feature $\widetilde{z}_m \in \mathbb{R}^{N \times T \times P \times D_l}$ by performing a Hadamard product of $h_m$ with each node dimension of $z_m$, \textit{i.e.},
\begin{equation}
    \vspace{-5pt}
    \widetilde{z}_m = \underbrace{\bracks{h_m; h_m; \dots}}_{D_l \textrm{ times}} \odot z_m.
    \label{eq:z_recons}
    \vspace{-5pt}
\end{equation}
In other words, we aim to pick the latent features in $z_m$ that correspond to the most representative frames in $\mathcal{X}_m$. While training, our scorer successfully learns to assign higher $h_m$ values to the $z_m$ features representing the more representative frames, and favors them in the reconstruction process.

From the weighted latent feature $\widetilde{z}_m$, our decoder produces a reconstruction $\widetilde{\mathcal{X}}_m \in \mathbb{R}^{N \times T \times P \times D}$ of the input graph nodes using another STGCN, \textit{i.e.},
\begin{equation}
    \vspace{-5pt}
    \widetilde{\mathcal{X}}_m = \textrm{STGCN}\parens{\mathcal{A}_m, \widetilde{z}_m; W_m^{\parens{\textrm{dec}}}},
    \label{eq:x_recons}
    \vspace{-5pt}
\end{equation}
where $W_m^{\parens{\textrm{dec}}}$ consists of the set of trainable STGCN parameters in the decoder.

\subsection{Loss Function for Training}\label{subsec:loss}
Analogous to Eq.~\ref{eq:net_difference}, we train our network architecture to maximally reconstruct the input graph nodes in all the modalities while minimizing the number of frames considered for reconstruction. Our approach is based on the assumption that the frames with higher representativeness, constitute the more highlightable excerpts of the video. Therefore, in effect, we aim to suppress as many frames as possible in the reconstruction of the input video while focusing on only the frames with high representativeness.

Given the highlight scores $h_m$ for each modality $m$, we perform a max-pooling of the scores across all dimensions but the time to obtain $h_m^{\parens{\textrm{max } T}} \in \bracks{0, 1}^{T \times 1}$, the maximum highlight score per frame of the video for that modality, \textit{i.e.},
\begin{equation}
    \vspace{-5pt}
    h_m^{\parens{\textrm{max } T}} = \max_{n \in N, p \in P} h_m.
    \vspace{-5pt}
\end{equation}

We also consider a weighted contribution of $h_m^{\parens{\textrm{max } T}}$ for each modality $m$, such that the weight is proportional to the number of frames in which the modality was visible in the input video. We define a modality to be observable in a frame if more than half the constituent points of that modality are visible in the frame. By that definition, we construct a weight $\alpha_m$ for each modality $m$ as
\begin{equation}
    \vspace{-5pt}
    \alpha_{m} = \frac{\textrm{\# frames where modality } m \textrm{ is observable}}{T}.
\end{equation}
We have $0 \leq \alpha_m \leq 1$ $\forall m$ since each frame can contain between no and all modalities.

We then construct weighted highlight scores $\bar{h}_m \in \bracks{0, 1}^{T \times 1}$ for all the frames of the video as
\begin{equation}
    \vspace{-5pt}
    \bar{h}_m = \alpha_m h_m^{\parens{\textrm{max } T}}.
    \label{eq:weighted_hlt_score}
    \vspace{-5pt}
\end{equation}

Finally, given the decoder reconstructions $\widetilde{\mathcal{X}}_m$ and the weights per modality $\alpha_m$, we construct our loss function $\mathcal{L}$ for training our network as
\begin{equation}
    \vspace{-5pt}
    \mathcal{L} = \sum_m \norm{\mathcal{X}_m - \widetilde{\mathcal{X}}_m} + \norm{\bar{h}_m} + \lambda_m\norm{W_m},
    \label{eq:loss_function}
    \vspace{-5pt}
\end{equation}
where $W_m$ collates all the trainable parameters $W_m^{\parens{\textrm{enc}}}$, $W_m^{\parens{\textrm{hlt}}}$, and $W_m^{\parens{\textrm{dec}}}$, $\lambda_m$ are the regularization factors, and we use the smooth-$\ell_1$ norm for $\norm{\cdot}$. We note that $\mathcal{L}$ consists of contrasting objectives that provide the competition needed to learn the highlight scores. The subtrahend $\widetilde{X}_m$ in the first term, $\norm{\mathcal{X}_m - \widetilde{\mathcal{X}}_m}$, obtained from Eqs.~\ref{eq:z_recons} and~\ref{eq:x_recons}, is a stand-in for the subtrahend in Eq.~\ref{eq:net_difference}. Minimizing this first term would require setting all highlight scores to $1$ (so all frames are highlights). Conversely, minimizing the second term $\norm{\bar{h}_m}$ would require setting all highlight scores to $0$ (so no frames are highlights). Consequently, our network ends up assigning high highlight scores to only the set of frames with maximal representativeness.

\section{Implementation and Testing}\label{sec:impl_and_test}
We train our network on the large-scale AVA-Kinetics dataset~\cite{ava_v2}. This dataset consists of $235$ training videos and $64$ validation videos, each $15$ minutes long and annotated with action labels in $1$-second clips. We ignore the action labels and use the original videos to train and validate our highlight detection network. The dataset consists of a wide variety of human activities but no supervision on highlightable excerpts. Thus, it is suitable for our task of learning to detect human-specific highlight excerpts. Owing to memory constraints, we process each video in non-overlapping excerpts of $30$ seconds, leading to a total of $7$,$050$ training excerpts and $1$,$920$ excerpts for validation.

\subsection{Implementation}\label{subsec:impl}
We use $M=2$ modalities, poses and faces, which are the two most observable modalities in all the datasets we tested our method on. Other modalities, such as hand gestures and eye movements, are either rarely visible or suffer from noisy detection. We build the pose graph following the CMU panoptic model~\cite{panoptic,3d_pose_detect}, and the face landmarks graph following the face landmarks model of Geitgey~\cite{face_landmark_detect}.

We use a multi-person tracker~\cite{mpt} to track the persons across all the frames. We use a pose detector~\cite{3d_pose_detect} and a face landmark detector~\cite{face_landmark_detect}, to respectively detect the coordinates of their 3D poses and 2D face landmarks. We scale all the coordinates to lie in the range $\bracks{-1, 1}$. To build our graph for each modality, we consider up to $P=20$ persons in each frame and temporal edges to $30f$ temporally adjacent frames combining the past and the future, $f$ being the frame rate of processing the videos. When available, we use an equal number of frames in the past and the future for temporal adjacency. We have observed efficient performance in terms of both accuracy and memory requirements for frame rates between $2$ and $5$, use $f=5$ for our experiments. For all $z_m$'s, we use a latent dimension of $D_l=8$.

We train using the Adam optimizer~\cite{adam} for $200$ epochs with a batch size of $2$, an initial learning rate of $10^{-3}$, a momentum of $0.9$, and a weight decay of $10^{-4}$. We decrease our learning rate by a factor of $0.999$ after every epoch. Our training took around $4.6$ GPU days at around $40$ minutes per epoch on an Nvidia GeForce GTX 1080Ti GPU.

\begin{table}[t]
    \centering
    \caption{Mean average precision on the DSH dataset~\cite{lsvm_dsh}. Bold: \textbf{best}, underline: \underline{second-best}. Our method performs second-best in the surfing domain, where not enough poses and faces were detected, and best in all the other domains.}
    \label{tab:mean_ap_dsh}
    \resizebox{\columnwidth}{!}{%
    \begin{tabular}{lC{1.1cm}C{1.6cm}C{1.1cm}C{1.5cm}C{1.1cm}}
    \toprule
    Domain & RRAE \cite{rrae} & Video2 GIF \cite{video2gif} & LSVM \cite{lsvm_dsh} & Less is More \cite{less_is_more} & Ours \\
    \midrule
    dog show & 0.49 & 0.31 & \underline{0.60} & 0.58 & \textbf{0.63} \\
    gymnastics & 0.35 & 0.34 & 0.41 & \underline{0.44} & \textbf{0.73} \\
    parkour & 0.50 & 0.54 & 0.61 & \underline{0.67} & \textbf{0.72} \\
    skating & 0.25 & 0.55 & \underline{0.62} & 0.58 & \textbf{0.64} \\
    skiing & 0.22 & 0.33 & 0.36 & \underline{0.49} & \textbf{0.52} \\
    surfing & 0.49 & 0.54 & 0.61 & \textbf{0.65} & \underline{0.62} \\
    \midrule
    Mean & 0.38 & 0.46 & 0.54 & \underline{0.57} & \textbf{0.64} \\
    \bottomrule
    \end{tabular}
    }
    \vspace{-5pt}
\end{table}

\subsection{Testing}\label{subsec:testing}
At test time, we obtain weighted highlight scores $\sum_m \bar{h}_m$ following Eq.~\ref{eq:weighted_hlt_score} for each frame of the input video. We combine all contiguous frames above a threshold $h_{\textrm{thres}}$ to generate highlight excerpts for the video. Based on our experiments, we have observed that values of $h_{\textrm{thres}} \geq 0.5$ leads to the detection of representative highlight excerpts in the benchmark datasets. The difference between $h_{\textrm{thres}}$ and $\tau$ (Section~\ref{subsec:representativeness}) is that $h_{\textrm{thres}}$ is used for trained highlight scores that capture domain- and user-preference-agnostic representativeness. In practice, we assign the individual highlight excerpts a score that is the mean of the weighted highlight scores for each of its constituent frames. We rank the excerpts based on these scores so that users can select their own thresholds to obtain the excerpts above those thresholds. The higher the threshold they choose, the fewer excerpts that survive the thresholding, thus reducing their manual effort of sifting through less representative excerpts.

\begin{table}[t]
    \centering
    \caption{Mean average precision on PHD$^2$~\cite{phd2}. Bold: \textbf{best}, underline: \underline{second-best}.}
    \label{tab:mean_ap_phd2}
    \resizebox{\columnwidth}{!}{%
    \begin{tabular}{C{1.5cm}C{1.5cm}C{1.5cm}C{1.5cm}C{1.5cm}}
    \toprule
    Random & FCSN \cite{fcsn} & Video2 GIF \cite{video2gif} & Ad-FCSN \cite{adaptive_fcsn} & Ours \\
    \midrule
    0.12 & 0.15 & 0.15 & \underline{0.16} & \textbf{0.20} \\
    \bottomrule
    \end{tabular}
    }
    \vspace{-15pt}
\end{table}

\section{Experiments}\label{sec:experiments}
We evaluate the comparative performance of our method and current state-of-the-art methods on two large-scale public benchmark datasets: the Domain-Specific Highlights (DSH) dataset~\cite{lsvm_dsh} and the Personal Highlight Detection dataset (PHD$^2$)~\cite{phd2}. We also evaluate on the smaller public datasets of TVSum~\cite{tvsum} and SumMe~\cite{summe}. Unlike any of the current approaches, however, we do not train or fine-tune our method on any of these datasets. We also test the performance of ablated versions of our network by removing individual modalities from training and evaluation.

\subsection{Datasets}\label{subsec:datasets}
The DSH dataset~\cite{lsvm_dsh} consists of YouTube videos across six domain-specific categories: dog show, gymnastics, parkour, skating, skiing, and surfing. There are roughly $100$ videos in each domain, with a total duration of around $1$,$430$ minutes. The PHD$^2$ dataset~\cite{phd2} consists of a total of around $10$,$000$ YouTube videos in the test set, totaling about $55$,$800$ minutes. It consists of highlights annotated by $850$ users based on their preferences. The TVSum dataset~\cite{tvsum} has $50$ YouTube videos totaling about $210$ minutes, collected across ten domains: beekeeping (BK), bike tricks (BT), dog show (DS), flash mob (FM), grooming animal (GA), making sandwich (MS), parade (PR), parkour (PK), vehicle tire (VT), and vehicle unstuck (VU). The SumMe dataset~\cite{summe} has $25$ personal videos, totaling about $66$ minutes.

\subsection{Evaluation Metrics}\label{subsec:eval_metrics}
We compute the commonly used mean average precision (mAP) of the detected highlights matching the annotated highlights~\cite{lsvm_dsh,video2gif,phd2,less_is_more,adaptive_fcsn}. For evaluating highlights, we consider the precision for each video individually rather than across videos, because the highlights detected from one video need not necessarily have higher highlight scores than the non-highlighted segments of another video~\cite{lsvm_dsh}. We also report the mean F-score (harmonic mean of the precision and the recall, calculated per video, and then averaged over all videos) of our method on all the datasets and for the provided baselines on the SumMe dataset~\cite{summe}.

\begin{table}[t]
    \centering
    \caption{Mean average precision on the TVSum dataset~\cite{tvsum}. Full domain names are in Section~\ref{subsec:datasets}. Bold: \textbf{best}, underline: \underline{second-best}. Our method performs second-best in the domains that are not fully human-centric (BK, DS, GA, MS), and best in all the other domains.}
    \label{tab:mean_ap_tvsum}
    \resizebox{\columnwidth}{!}{%
    \begin{tabular}{lC{1.0cm}C{1.0cm}C{1.0cm}C{1.8cm}C{1.5cm}C{1.0cm}}
    \toprule
    Domain & MBF \cite{mbf} & KVS \cite{category_specific_summ} & CVS \cite{collab_summ} & Adv-LSTM \cite{unsup_video_summ} & Less is More \cite{less_is_more} & Ours \\
    \midrule
    BK & 0.31 & 0.34 & 0.33 & 0.42 & \textbf{0.66} & \underline{0.57} \\
    BT & 0.37 & 0.42 & 0.40 & 0.48 & \underline{0.69} & \textbf{0.93} \\
    DS & 0.36 & 0.39 & 0.38 & 0.47 & \textbf{0.63} & \underline{0.60} \\
    FM & 0.37 & 0.40 & 0.37 & \underline{0.46} & 0.43 & \textbf{0.88} \\
    GA & 0.33 & 0.40 & 0.38 & 0.48 & \textbf{0.61} & \underline{0.50} \\
    MS & 0.41 & 0.42 & 0.40 & 0.49 & \textbf{0.54} & \underline{0.50} \\
    PR & 0.33 & 0.40 & 0.38 & 0.47 & \underline{0.53} & \textbf{0.84} \\
    PK & 0.32 & 0.38 & 0.35 & 0.46 & \underline{0.60} & \textbf{0.76} \\
    VT & 0.30 & 0.35 & 0.33 & 0.42 & \underline{0.56} & \textbf{0.65} \\
    VU & 0.36 & 0.44 & 0.41 & 0.47 & \underline{0.50} & \textbf{0.77} \\
    \midrule
    Mean & 0.35 & 0.40 & 0.37 & 0.46 & \underline{0.58} & \textbf{0.70} \\
    \bottomrule
    \end{tabular}
    }
    \vspace{-5pt}
\end{table}

\begin{table}[t]
    \centering
    \caption{F-scores on the SumMe dataset~\cite{summe}. Bold: \textbf{best}, underline: \underline{second-best}.}
    \label{tab:f_score_summe}
    \resizebox{\columnwidth}{!}{%
    \begin{tabular}{C{1.2cm}C{1.1cm}C{1.1cm}C{1.1cm}C{1.1cm}C{1.1cm}C{1.1cm}C{1.1cm}}
    \toprule
    Int \cite{summe} & Sub \cite{submodular_mixtures_1} & DPP-LSTM \cite{sup_video_summ} & GAN-S \cite{unsup_video_summ} & DRL-S \cite{drl_summ} & S$^2$N \cite{s2n} & Ad-FCSN \cite{adaptive_fcsn} & Ours \\
    \midrule
    0.39 & 0.40 & 0.39 & 0.42 & 0.42 & 0.43 & \underline{0.44} & \textbf{0.48} \\
    \bottomrule
    \end{tabular}
    }
    \vspace{-15pt}
\end{table}

\subsection{Baselines}\label{subsec:baselines}
We compare with four baselines on the DSH dataset~\cite{lsvm_dsh}, four on PHD$^2$~\cite{phd2}, five on the TVSum dataset~\cite{tvsum}, and seven on the SumMe dataset~\cite{summe}. We report the performances of the baselines as stated in the literature.

On the DSH dataset, we compare with the latent SVM-based highlight ranking (LSVM) method of Sun et al.~\cite{lsvm_dsh}, Video2GIF~\cite{video2gif}, which uses C3D features with fully connected layers to learn highlight ranking, the unsupervised robust recurrent autoencoder method (RRAE) of Yang et al.~\cite{rrae}, and the method of Xiong et al. (Less is More)~\cite{less_is_more} that learns to rank highlights by using the duration of videos as weak supervision, with the insight that shorter videos are more likely to be edited and therefore more highlightable.

On PHD$^2$, we compare with Video2GIF~\cite{video2gif} again, the fully convolutional sequence network (FCSN) that uses GoogLeNet to learn image-based features for highlight detection~\cite{fcsn}, and the adaptive FCSN method (Ad-FCSN)~\cite{adaptive_fcsn}, which additionally consists of a history encoder to adapt to a user's history of highlight preferences to detect personalized highlights. We also use a fully random highlight detector as the lowest baseline following~\cite{adaptive_fcsn}.

On the TVSum dataset, we compare again with the duration-based highlight detection method (Less is More)~\cite{less_is_more}, the visual correlation-based method of Chu et al.~\cite{mbf} that uses maximal biclique finding (MBF) to obtain co-occurring shots that are also relevant to the original video, the kernel-based video summarization method (KVS) of Potapov et al.~\cite{category_specific_summ} that trains an SVM on semantically consistent segments, the collaborative video summarization method (CVS) of Panda et al.~\cite{collab_summ} that uses a consensus regularizer to detect highlight segments satisfying sparsity, diversity, and representativeness, and the unsupervised video summarization method of Mahasseni et al.~\cite{unsup_video_summ} using LSTMs with adversarial loss (Adv-LSTM).

On the SumMe dataset, we compare again with adaptive FCSN (Ad-FCSN)~\cite{adaptive_fcsn}, the interestingness-based summarization method (Int.) of Gygli et al.~\cite{summe}, the submodularity-based summarization method (Sub.) of Gygli et al.~\cite{submodular_mixtures_1}, the LSTM network of Zhang et al.~\cite{sup_video_summ} employing a determinantal point process (DPP-LSTM), the GAN-based method of Lu and Grauman~\cite{story_driven_summ} with extra supervision (GAN-S), the deep reinforcement learning-based method of Zhou et al.~\cite{drl_summ} with extra supervision (DRL-S), and the sequence to segments detection method (S$^2$N)~\cite{s2n} that uses an encoder-decoder architecture to detect segments with high relevance from sequence data.

\begin{table}[t]
    \centering
    \caption{Comparison of mean mAP and mean F-score for different ablated versions of our method on benchmark datasets. Bold: \textbf{best}, underline: \underline{second-best}.}
    \label{tab:ablation}
    \resizebox{0.9\columnwidth}{!}{%
    \begin{tabular}{lcccccc}
        \toprule
        \multirow{2}{*}{Dataset} & \multicolumn{6}{c}{Using Modality} \\
        \cmidrule{2-7}
        & \multicolumn{2}{c}{Face only} & \multicolumn{2}{c}{Pose only} & \multicolumn{2}{c}{Both} \\
        \cmidrule{2-7}
        & mAP & F & mAP & F & mAP & F \\
        \midrule
        DSH~\cite{lsvm_dsh} & 0.51 & 0.45 & \underline{0.57} & \underline{0.48} & \textbf{0.64} & \textbf{0.56} \\
        TVSum~\cite{tvsum} & 0.57 & 0.46 & \underline{0.64} & \underline{0.56} & \textbf{0.70} & \textbf{0.59} \\
        PHD$^2$~\cite{phd2} & \underline{0.16} & \underline{0.20} & 0.15 & 0.18 & \textbf{0.20} & \textbf{0.22} \\
        SumMe~\cite{summe} & \underline{0.48} & 0.39 & 0.45 & \underline{0.41} & \textbf{0.52} & \textbf{0.48} \\
        \bottomrule
    \end{tabular}
    }
    \vspace{-5pt}
\end{table}

\begin{figure}[t]
    \centering
    \includegraphics[width=0.7\columnwidth]{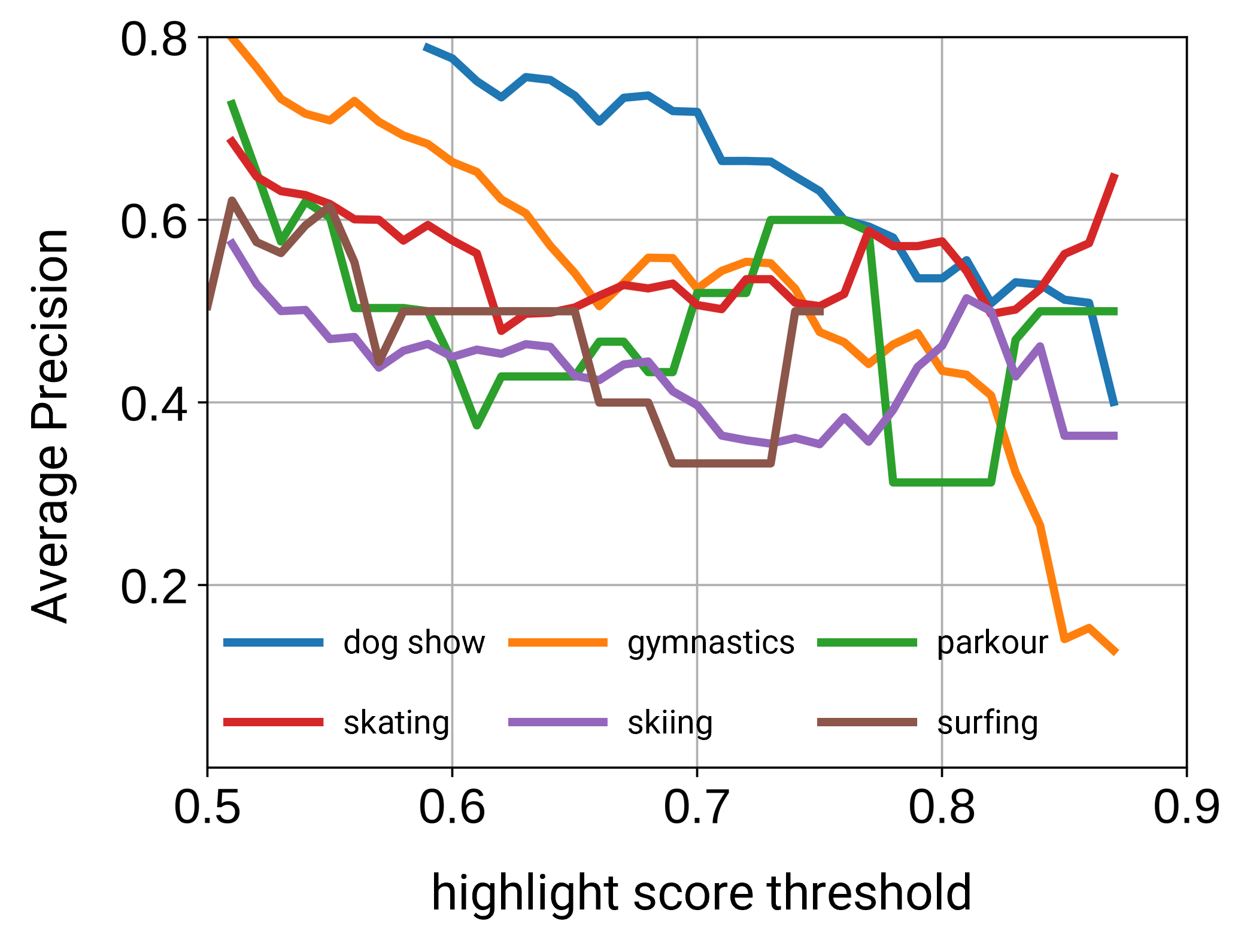}
    \caption{\textbf{Average precision by highlight score threshold $h_{\textrm{thres}}$.} On the domains in the DSH dataset~\cite{lsvm_dsh}.}
    \label{fig:ap_by_thres}
    \vspace{-15pt}
\end{figure}

\begin{figure}[t]
    \centering
    \includegraphics[width=0.9\columnwidth]{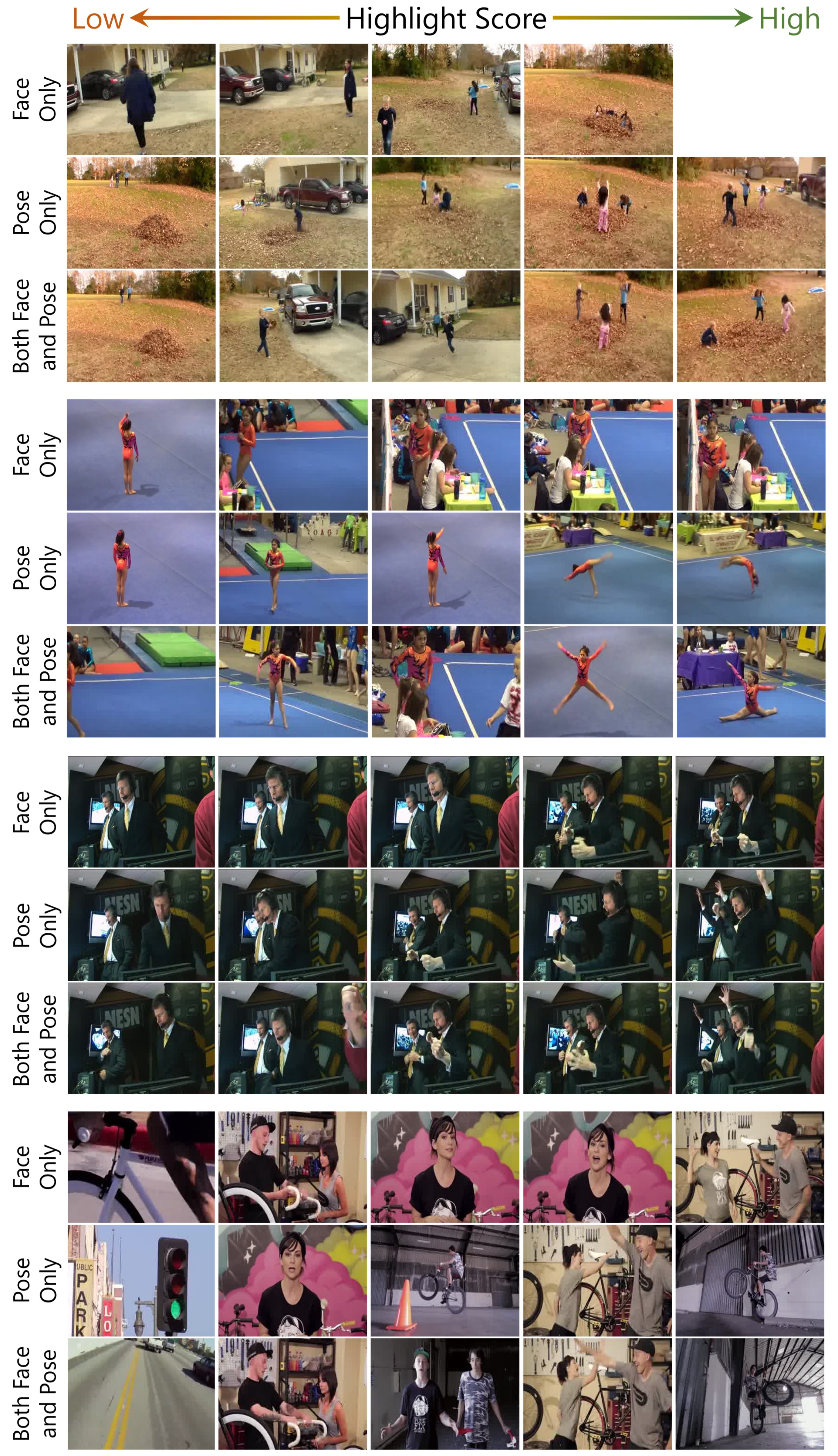}
    \caption{\textbf{Sample highlight frames detected by our method.} We show sample frames across the range of highlight scores as detected by different ablated versions of our method. We show one sample video from the datasets SumMe~\cite{summe}, PHD$^2$~\cite{phd2}, DSH~\cite{lsvm_dsh}, and TVSum~\cite{tvsum}, in order from top to bottom. When using only faces or only poses, our method learns highlight scores based only on face- or pose-based representativeness. Combining both the modalities, our method learns highlight scores based on representativeness from both.}
    \label{fig:qualitative}
    \vspace{-15pt}
\end{figure}

\subsection{Results}
\textbf{DSH~\cite{lsvm_dsh} and TVSum~\cite{tvsum}.} We report the mAP across all the domains in these datasets in Tables~\ref{tab:mean_ap_dsh} and~\ref{tab:mean_ap_tvsum} respectively. We outperform the baselines on all but a few domains, which are either not fully human centric (beekeeping, dog show, grooming animals, and making sandwich in TVSum), or where sufficient poses and faces could not be detected (surfing in DSH). However, we come second-best on these domains, and on average, across the domains, we outperform the best baselines by an absolute $4\%-12\%$.

\textbf{PHD$^2$~\cite{phd2}}. We report the mAP across the dataset in Table~\ref{tab:mean_ap_phd2}. Given the abundance of humans detected in the videos, we outperform the best baseline by an absolute $4\%$.

\textbf{SumMe~\cite{summe}.} We report the mean F-scores across the dataset in Table~\ref{tab:f_score_summe}. Following prior methods~\cite{s2n,adaptive_fcsn}, we randomly select $20\%$ of the dataset for calculating the mean F-score, repeat this experiment five times, and report the mean performance. Based on these experiments, we outperform the best baseline by an absolute $4\%$.

These results demonstrate that our approach of using human-centric modalities to detect highlights leads to state-of-the-art performance on all these benchmark datasets.

\subsection{Ablation Studies}\label{subsec:ablation}
In our experiments, we consider two modalities, poses and faces. We ablate each of these two modalities in turn and test the performance of our method by training it on the remaining modality. We report the mean mAP and the mean F-score of the ablated versions of our method on all four benchmark datasets in Table~\ref{tab:ablation}. Using only poses and no faces, we observe an absolute drop-off of $5\%-7\%$ for the mean mAP and $3\%-8\%$ for the mean F-score across the datasets, compared to using both modalities. Using only faces and no poses, we observe more severe absolute drop-offs of $4\%-13\%$ for the mean mAP and $2\%-13\%$ for the mean F-score across the datasets. This happens because poses are generally more abundant and more easily detected compared to face landmarks. For example, poses can be detected even when a human is partially occluded, in the dark, or not in clear focus, whereas detection of face landmarks requires the face to be well-lit and in focus. Therefore, not detecting poses leads to missing a significant number of highlightable excerpts. This trend is reversed only in PHD$^2$, where faces were more commonly detected than poses.

We also show the qualitative performance of our method and all its ablated versions on one sample video from each of the four datasets, DSH, PHD$^2$, TVSum, and SumMe, in Figure~\ref{fig:qualitative}. We see that when observing only poses and not faces, our method detects the representative highlight excerpts with pose-based expressions but fails to detect excerpts that primarily have facial expressions and emotions. Conversely, when observing only faces and not poses, our method can only detect the excerpts where the faces are prominent, and misses excerpts where the faces are too small, occluded, or in the dark. Using both modalities, our method can detect all the representative excerpts.

\subsection{Effect of Highlight Score Threshold}

We use a threshold $h_{\textrm{thres}}$ on the highlight score predicted by our method to detect the highlightable excerpts (Section~\ref{subsec:testing}). To visualize the effect $h_{\textrm{thres}}$ has on the average precision (AP), we show the plot of AP vs. $h_{\textrm{thres}}$ on each domain in the DSH dataset~\cite{lsvm_dsh} in Figure~\ref{fig:ap_by_thres}. We observe a general trend of the AP decreasing as we increase the threshold, as our method returns fewer and fewer highlights. However, this is not true for some domains like surfing, where the highlight scores of the representative excerpts are already sufficiently high. In practice, we consider the choice of threshold to be user-configurable for each video.

\section{Conclusion, Limitations and Future Work}\label{sec:conclusion}
We have presented a novel method for detecting highlights from human-centric videos, which leverages the observable human-centric modalities, such as faces and poses, and uses these modalities to automatically detect the most representative highlights from the video. Extensive experimental results on the domain-specific highlights (DSH) dataset~\cite{lsvm_dsh}, the personal highlight detection dataset (PHD$^2$)~\cite{phd2}, the TVSum dataset~\cite{tvsum}, and the SumMe dataset~\cite{summe} demonstrate the benefits of our proposed approach compared to several state-of-the-art baselines.

Our work has some limitations. Although our network design can accommodate any number of modalities, we have used only faces and poses as the modalities in our benchmarks. This leads to state-of-the-art results on average. However, many videos (\textit{e.g.}, videos on grooming animals, making sandwich in TVSum) exhibit other modalities such as hands and fingers. Thus, we plan to incorporate more human-centric modalities into our experiments in the future. Our method may not offer much improvements in videos from domains that are not fully human-centric, \textit{e.g.}, videos focusing on other living beings, inanimate objects, and natural scenes. We plan to explore these domains in the future using appropriate modalities. Our method can also be combined with domain-specific features and adapted to user preferences to detect more fine-tuned highlights.

\section*{Acknowledgment}
This work was supported in part by ARO Grants W911NF1910069 and W911NF2110026.

\clearpage

{\small
\bibliographystyle{ieee_fullname}
\bibliography{main}
}

\end{document}